\documentclass[a4paper, 10pt, conference]{ieeeconf}      

\IEEEoverridecommandlockouts                              

\usepackage[table]{xcolor}
\usepackage{amsmath,amssymb,amsfonts}
\usepackage{algorithm}
\usepackage{array}
\usepackage{textcomp}
\usepackage{stfloats}
\usepackage{url}
\usepackage{verbatim}
\usepackage{graphicx}
\usepackage{cite}
\usepackage{subfiles}   
\usepackage[hidelinks]{hyperref}
\usepackage{amsfonts}
\usepackage{booktabs}
\usepackage{multirow}
\usepackage{pifont}
\usepackage{float}
\usepackage{algpseudocode}
\usepackage{siunitx}
\usepackage[framemethod=TikZ]{mdframed}
\usepackage{xcolor}
\usepackage{pgf}
\usepackage{tikz}
\usepackage{tikzscale}
\usepackage{pgfplots}
\pgfplotsset{compat=1.18}
\usepackage{pgf}
\usepackage{graphicx}
\usepackage{tikz}
\usepackage{textcomp}
\usepackage{amsmath}
\usetikzlibrary{fit}

\usepackage{tikz}
\usepackage{tikz-cd}
\usepackage{pgfplots}
\pgfplotsset{compat=1.14}

\definecolor{custompink}{RGB}{255, 105, 180} 

\title{\LARGE \bf
End-to-End Multi-Task Policy Learning from NMPC for Quadruped Locomotion
}

\author{Anudeep Sajja$^{1}$, Shahram Khorshidi$^2$, Sebastian Houben$^{1,3}$, Maren Bennewitz$^2$
\thanks{$^{1}$Authors are with the Institute for Artificial Intelligence and Autonomous Systems (A2S) at the Bonn-Rhein-Sieg University of Applied Sciences, Sankt Augustin, Germany.}%
\thanks{$^{2}$Authors are with the Humanoid Robots Lab, University of Bonn,
Germany, and additionally with the Lamarr Institute for Machine Learning
and Artificial Intelligence and the Center for Robotics, Bonn, Germany.}%
\thanks{$^{3}$Author is with the Fraunhofer Institute for Intelligent Analysis and Information Systems, Sankt Augustin, Germany.}%
\thanks{*This work has been funded by the Federal Ministry of Education and Research of Germany and the state of North-Rhine Westphalia as part of the Lamarr Institute for Machine Learning and Artificial Intelligence, LAMARR22B.}
}

\begin{document}

\maketitle
\thispagestyle{empty}
\pagestyle{empty}

\begin{abstract}
Quadruped robots excel in traversing complex, unstructured environments where wheeled robots often fail. However, enabling efficient and adaptable locomotion remains challenging due to the quadrupeds' nonlinear dynamics, high degrees of freedom, and the computational demands of real-time control. Optimization-based controllers, such as Nonlinear Model Predictive Control (NMPC), have shown strong performance, but their reliance on accurate state estimation and high computational overhead makes deployment in real-world settings challenging. In this work, we present a Multi-Task Learning (MTL) framework in which expert NMPC demonstrations are used to train a single neural network to predict actions for multiple locomotion behaviors directly from raw proprioceptive sensor inputs. We evaluate our approach extensively on the quadruped robot Go1, both in simulation and on real hardware, demonstrating that it accurately reproduces expert behavior, allows smooth gait switching, and simplifies the control pipeline for real-time deployment. Our MTL architecture enables learning diverse gaits within a unified policy, achieving high \( R^2 \) scores for predicted joint targets across all tasks. The results of this research can be found   
\textcolor{custompink}{\href{https://youtu.be/zkpSi7tNKL4}{here}}.
\end{abstract}

\section{Introduction}\label{sec:introduction}
Quadruped robots are increasingly deployed in real-world applications such as search and rescue, firefighting, and service robotics, owing to their ability to traverse complex and unstructured environments. Their adaptability provides a significant advantage over wheeled robots, particularly in rough or natural terrains where conventional platforms often fail.

Despite significant progress, achieving agile and robust locomotion in quadruped robots remains a challenging problem \cite{ha2024learningbasedleggedlocomotionstate, wensing2023optimization}. Early methods relied on simplified models and heuristic strategies \cite{heursticbased},
\begin{figure}[!ht]
    \centering
    \includegraphics[width=1.0\linewidth]{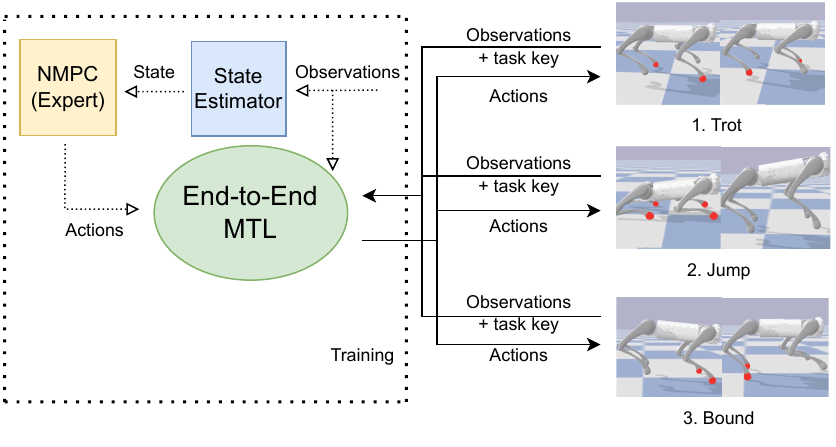}
    \vspace{-0.95cm}
    \caption{Overview of our end-to-end multi-task learning framework for quadrupedal locomotion. A single neural network learns multiple gait-specific policies from an expert NMPC via imitation learning. The shared policy maps proprioceptive observations to joint targets. Red markers indicate foot contact points, highlighting the contact sequence for each gait.}
    \vspace{-0.6cm}
    \label{fig:end2end_framework_go1}
\end{figure}
but these approaches struggled to handle complex terrain or rapidly changing locomotion patterns. This led to the development of biologically inspired approaches, such as central pattern generators \cite{cpg}, enabling rhythmic and more natural gait generation. The field further advanced with the adoption of Optimal Control (OC) \cite{wensing2023optimization}, which offers a principled framework for planning motion trajectories while explicitly handling system constraints and optimizing performance cost. A real-time variant of OC, known as Model Predictive Control (MPC) \cite{Rawlings2017mpc_book}, solves finite-horizon control problems in a receding-horizon fashion and has proven effective in legged robotics locomotion. More recently, Nonlinear MPC (NMPC) \cite{Mastalli2022mpc, Grandia2023mpc} has demonstrated impressive results in generating agile and adaptive locomotion. However, NMPC remains computationally intensive, sensitive to local minima, and heavily dependent on accurate state estimation, making deployment on hardware challenging.

To address the drawbacks of NMPC, learning-based approaches have gained increasing attention. These methods aim to learn control policies that directly map sensory inputs to actions. Among these, Reinforcement Learning (RL) has shown promise for acquiring complex and adaptive locomotion behaviors through trial and error \cite{lee2020learning}. However, RL often suffers from high sample complexity and long training times, making real-world deployment difficult. In contrast, Imitation Learning (IL) offers a more sample-efficient alternative by learning from expert demonstrations \cite{khadiv2023learning}, avoiding the need for handcrafted reward functions and extensive exploration.

In this work, we leverage IL by using an NMPC expert to generate high-quality demonstration trajectories, which are then used to train a neural network policy that imitates the expert’s behavior. This combines the precision of optimization-based control with the efficiency of learning-based methods. A schematic overview of our approach is shown in Fig.~\ref{fig:end2end_framework_go1}.

Our key contributions are as follows:
\begin{itemize}
    \item We propose a Multi-Task Learning (MTL) framework that trains a single neural network to perform multiple quadrupedal gaits from expert demonstrations.
    \item The resulting policy is end-to-end, mapping raw proprioceptive sensor inputs directly to joint position targets, removing the need for intermediate state estimation.
    \item We evaluate our approach extensively in simulation and demonstrate its successful deployment on a real quadruped robot.
\end{itemize}

\section{Related Work}
Over the past decades, quadrupedal locomotion has been extensively studied using the OC framework and its real-time implementation via MPC \cite{Mastalli2022mpc, Grandia2023mpc, Meduri2023biconmp}. More recently, breakthroughs in learning-based methods have opened new possibilities for enhancing the adaptability and robustness of locomotion over diverse terrains and environments. Reinforcement Learning (RL) and Imitation Learning (IL) have emerged as two dominant approaches for learning locomotion policies. In particular, RL has demonstrated the ability to learn complex behaviors from scratch through trial and error, as seen in \cite{peng2017learning} and \cite{hwangbo2019learning}. However, while RL provides a powerful framework for general policy learning, it often suffers from high sample complexity, long training times, and the need for extensive simulation or real-world interactions, making real-time deployment and generalization across tasks challenging.

In contrast, IL offers a more sample-efficient alternative by learning directly from expert demonstrations. This has proven particularly effective when expert trajectories are available, either from human input or classical controllers. Chen et al.\cite{chen2020learning_by_cheating} used a student-teacher framework for vision-based agents, while Yang et al.\cite{Yang_lfd_semantics} trained quadrupeds to adapt gait and speed using joystick inputs. More relevant to our work, recent studies have used NMPC to generate high-quality expert data. For instance, Khadiv et al.\cite{khadiv2023learning} demonstrated that expert NMPC trajectories can be used to train dynamic gait policies through IL.

Another key challenge in learning locomotion policies is the choice of observation space. Early approaches typically relied on accurate state estimation modules to provide input to the control policy, introducing an additional layer of complexity and potential sources of error. To mitigate this, some works have explored learning dynamics models to improve state estimation in optimization-based frameworks \cite{khorshidi2023centroid, khorshidi2024koopman}, with physically consistent models shown to generalize better across tasks for estimation and control \cite{khorshidi25icra}. More recent IL and RL approaches have successfully learned policies directly from raw proprioceptive data, such as Inertial Measurement Unit (IMU) and joint encoder data \cite{lee2020learning, agarwal2023legged, ji_legged_locomotion}, eliminating the need for explicit state reconstruction and enabling simpler, more reactive controllers. However, most of these methods focus on single-task learning, requiring a separate network for each behavior—limiting scalability and adaptability.

Multi-task learning has emerged as a promising solution to this limitation by enabling a single model to learn multiple behaviors simultaneously \cite{Zhang_mtl_survey}. While MTL has shown strong results in domains such as computer vision and natural language processing, its application in legged robotics locomotion remains relatively unexplored. Most prior work continues to rely on per-task models or requires extensive retraining to switch between gaits \cite{khadiv2023learning}. Integrating MTL with IL is particularly appealing in the context of robotic control, where expert demonstrations are available, as it enables efficient reuse of shared representations across tasks and can simplify deployment in practice.

Building on these insights, we propose a single, end-to-end control policy that executes multiple quadrupedal gaits directly from raw proprioceptive inputs. Our MTL framework leverages NMPC-generated demonstrations to train a unified neural network capable of switching between gaits without requiring additional modules for estimation. This simplifies the control pipeline compared to prior work, which often relies on separate policies or multi-stage architectures. We validate the effectiveness of our method in both simulation and real-world experiments using the Unitree Go1 quadruped.

\section{Multi-Task Learning} \label{sec:methodology}
\subsection{Problem Statement}
We aim to learn a unified, end-to-end control policy for quadrupedal locomotion that can perform multiple gait behaviors using a single neural network. Let \( \mathcal{G} = \{ g_1, g_2, \dots, g_K \} \) denote a set of \( K \) locomotion behaviors (e.g., trot, bound, jump, walk). For each behavior \( g_k \in \mathcal{G} \), we collect a dataset \( \mathcal{D}_k = \{ (o_t, a_t) \}_{t=1}^{N_k} \), where \( o_t \in \mathcal{O} \) are raw proprioceptive observations recorded from the robot (e.g., joint positions, velocities, IMU signals), and \( a_t \in \mathcal{A} \) are the corresponding actions generated by the NMPC expert (e.g., desired joint positions).

Our objective is to train a single policy
\begin{align}
    \pi_\theta : \mathcal{O} \times \mathcal{G} \rightarrow \mathcal{A},
\end{align}
parameterized by \( \theta \), that maps observations \( o_t \) and a task identifier \( g_k \) to actions \( a_t \), such that
\begin{align}
    \pi_\theta(o_t, g_k) \approx a_t \quad \text{for all } (o_t, a_t) \in \mathcal{D}_k.
\end{align}

We formulate this as a supervised learning problem using behavior cloning. This framework allows the network to jointly learn multiple gaits within a single model, without relying on separate policies or intermediate state estimation. This approach enables a compact control architecture that performs multiple locomotion behaviors using a single neural network, conditioned on task identifier and driven only by raw proprioceptive inputs.

\subsection{Action Space}
We adopt a Proportional-Derivative (PD) control formulation to define a structured and interpretable action space, where the policy predicts desired joint positions that are transformed into torques via a PD controller. This approach is widely used in learning-based locomotion control \cite{hwangbo2019learning, agarwal2023legged}, and has been shown to improve sample efficiency and training stability \cite{peng2017learning}.

The relationship between the policy output and the applied joint torque is defined through the following PD control equation
\begin{equation}
    \tau = K_p \left(\pi_\theta(o_t, g_k) - q_t \right) + K_d \left(0 - v_t \right),
    \label{eq:pd_control}
\end{equation}
where \( q_t \) and \( v_t \) are the current joint positions and velocities, and \( \pi_\theta(o_t, g_k) \) is the network-predicted desired joint position for behavior \( g_k \).

This formulation offloads low-level error correction to the PD controller, allowing the network to focus on higher-level gait generation. Since the NMPC expert provides joint torque commands \( \tau \), we derive training targets for the network by inverting the PD control law:

\begin{equation}
    a_{t} = q_t + \frac{\tau + K_d v_t}{K_p},
    \label{eq:inverse_pd}
\end{equation}

This provides a consistent and tractable way to generate supervised learning targets from expert demonstrations.

\subsection{Multi-Task Neural Network Architecture}

Our control policy is implemented as an end-to-end feedforward neural network that maps raw proprioceptive inputs directly to joint position targets. The input vector \( o_t \in \mathbb{R}^{34} \) includes IMU measurements, joint positions, joint velocities, and foot contact indicators. The output is a 12-dimensional vector representing the desired joint positions for the current control step.

\begin{figure}[!ht]
        \centering
        \scalebox{0.85}{  
        \begin{tikzpicture}[node distance=1.5cm]
        
            \tikzstyle{input neuron}=[circle, draw=black, fill=blue!20, minimum size=8mm]
            \tikzstyle{hidden neuron}=[circle, draw=black, fill=green!30, minimum size=8mm]
            \tikzstyle{output neuron}=[circle, draw=black, fill=red!30, minimum size=8mm]
            \tikzstyle{connector}=[->, thick]
        
            \node[input neuron] (Input) at (0, 0) {};
            \node at (-1.5, 0) {Input};
        
            \node[hidden neuron] (Hidden1) at (0, -1.5) {2560n};
            \node at (-1.5, -1.5) {Hidden 1};
        
            \node[hidden neuron] (Hidden2) at (0, -3) {2560n};
            \node at (-1.5, -3) {Hidden 2};
        
            \node[hidden neuron] (Hidden3a) at (-2, -4.5) {2560n};
            \node at (-3.0, -5.3) {Hidden 3a};
        
            \node[hidden neuron] (Hidden3b) at (0, -4.5) {2560n};
            \node at (0.9, -5.3) {Hidden 3b};
        
            \node[hidden neuron] (Hidden3c) at (2, -4.5) {2560n};
            \node at (3.0, -5.3) {Hidden 3c};

            \node[output neuron] (OutputA) at (-2, -6.5) {};
            \node at (-3.0, -6.5) {Trot};
        
            \node[output neuron] (OutputB) at (0, -6.5) {};
            \node at (1.0, -6.5) {Jump};
        
            \node[output neuron] (OutputC) at (2, -6.5) {};
            \node at (3.2, -6.5) {Bound};

            \draw[connector] (Input) -- (Hidden1);
            \draw[connector] (Hidden1) -- (Hidden2);
        
            \draw[connector] (Hidden2) -- (Hidden3a);
            \draw[connector] (Hidden2) -- (Hidden3b);
            \draw[connector] (Hidden2) -- (Hidden3c);
        
            \draw[connector] (Hidden3a) -- (OutputA);
            \draw[connector] (Hidden3b) -- (OutputB);
            \draw[connector] (Hidden3c) -- (OutputC);
        
            \node[draw=black, dashed, fit=(Hidden1) (Hidden2)] (trunk) {};
            \node at (2.0, -2.25) {Shared Trunk};
        
        \end{tikzpicture}
        }
        \caption{Multi-Task End-to-End Neural Network Architecture. Raw proprioceptive inputs are processed through two shared hidden layers to learn common locomotion features, which are then branched into three task-specific heads—trot, jump, and bound, each predicting desired joint position targets.}
        \vspace{-0.5cm}
        \label{fig:MTL_NNArch}
    \end{figure}
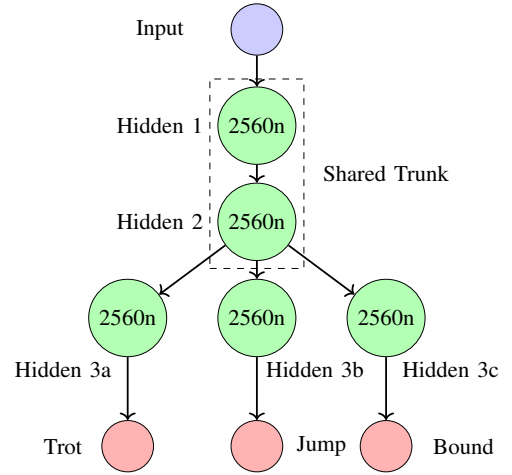

To enable multi-gait learning within a single network, we adopt a hard parameter sharing approach for MTL \cite{ruderMTL_overview}. As shown in Fig.~\ref{fig:MTL_NNArch}, the architecture consists of a shared trunk with two hidden layers, followed by task-specific output heads—one for each locomotion behavior (e.g., trot, bound, jump, walk). The shared layers capture general locomotion dynamics, while each head specializes in predicting joint targets specific to its corresponding gait. We used the Exponential Linear Unit (ELU) activation function \cite{clevert2015fast} to improve training speed and stability. All inputs were normalized following the procedure described in \cite{rutecki2020normalize}, ensuring consistent feature scaling.

Each data point is labeled with its corresponding behavior \( g_k \). The total loss is computed as the sum of mean squared errors across all tasks:
\begin{equation}
    \mathcal{L}_{\text{total}} = \sum_{k=1}^{K} \sum_{(o_t, a_{\text{t}}) \in \mathcal{D}_k} \left\| \pi_\theta(o_t, g_k) - a_{\text{t}} \right\|^2.
\end{equation}

During training, each task-specific head computes its own prediction loss based on the assigned gait label. These losses are summed into a single total loss, which is back-propagated through both the task-specific and shared layers. This structure allows the shared trunk to learn general locomotion features common across tasks, while the heads specialize in task-specific control outputs. As a result, the network benefits from improved data efficiency and the ability to learn multiple gaits within a single model.

\section{Experimental Results}\label{sec:results}
In this section, we evaluate our end‐to‐end multi-task imitation learning framework both in simulation and on the real hardware using the Unitree Go1 quadruped robot. In the simulation, we assess the ability of the learned policies to reproduce expert NMPC behaviors across multiple gaits—trot, bound, and jump and compare the performance of our MTL approach against a baseline single-task model. On the real robot, we evaluate the MTL policy by executing two gaits—trot and walk—using expert demonstrations collected directly from hardware experiments. We evaluate model performance using validation loss curves, report predictive accuracy using standard metrics (MSE, MAE, \( R^2 \)), and visualize selected joint target angles. For visualization, we focus on the front-left leg’s hip, thigh, and knee joints to illustrate how closely the predicted trajectories align with those of the NMPC expert.

\subsection{Simulation Experiments}
    \subsubsection{PyBullet Simulation Setup}
    All simulation experiments are conducted in the PyBullet physics simulator \cite{coumans2016pybullet}, using a training setup informed by the NMPC formulation of Meduri et al. \cite{Meduri2023biconmp} and state estimation insights from Agarwal et al.~\cite{AgarwalPybullet_state_est}.
    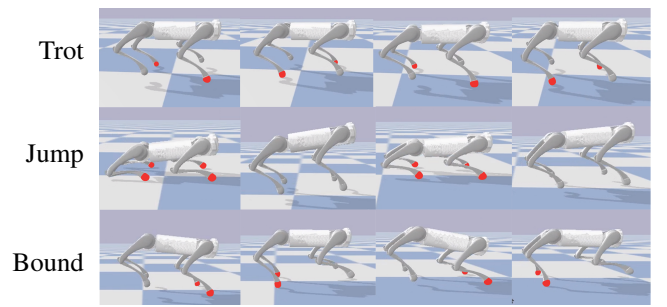
\begin{figure}[b]
         \centering
         \begin{tikzpicture}
             \node[inner sep=2pt] (img) at (0,0) {\includegraphics[width=0.85\linewidth]{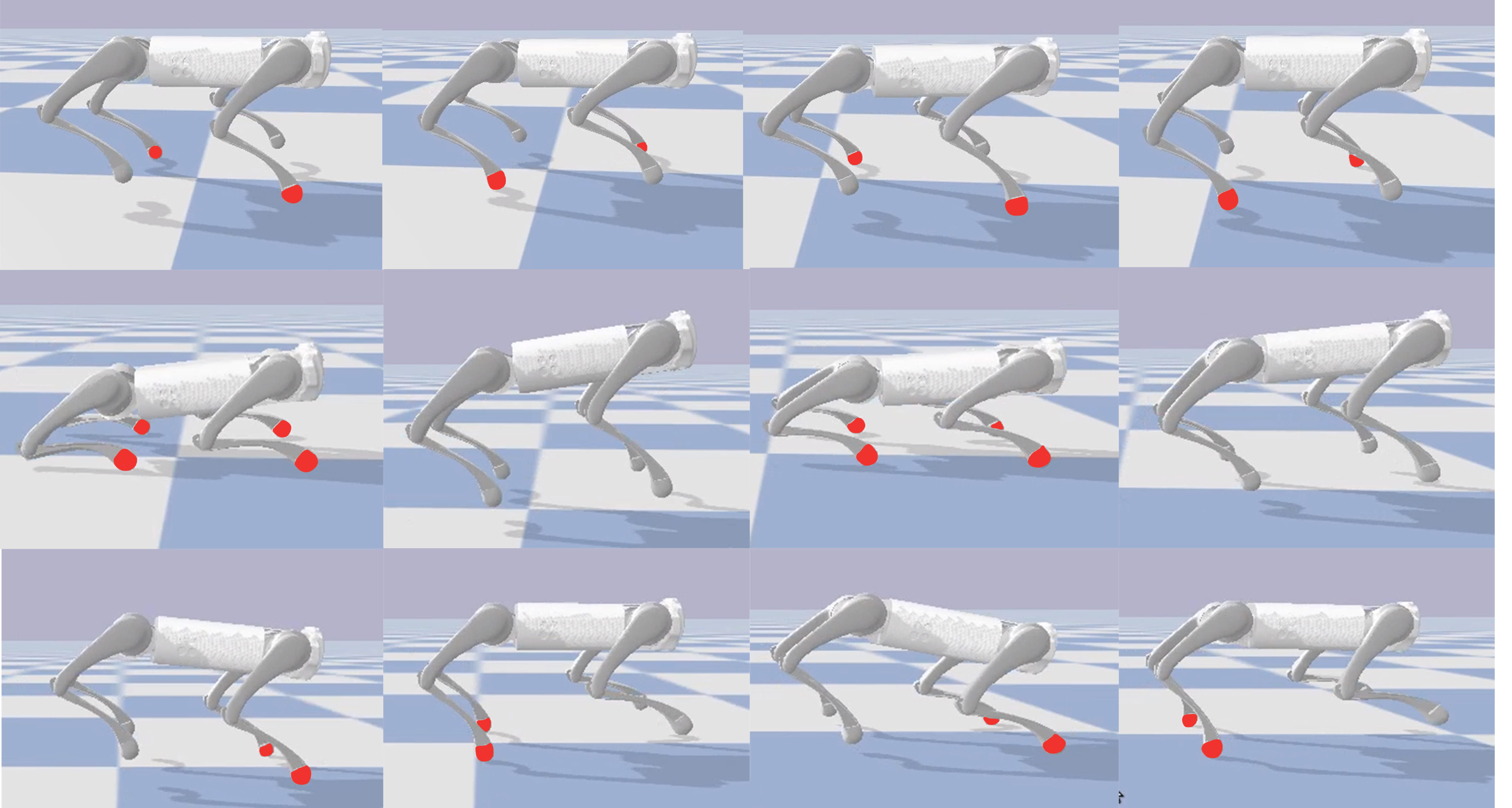}};
            \node[anchor=east] at ([yshift=40pt]img.west) {Trot};
            \node[anchor=east] at ([yshift=0pt]img.west) {Jump};
            \node[anchor=east] at ([yshift=-40pt]img.west) {Bound};
        \end{tikzpicture}
         \caption{Simulation snapshots of NMPC expert behaviors in PyBullet. Red markers indicate foot contact points, highlighting the contact sequence for each gait.}
         \label{fig:sim_image}
    \end{figure}
    
    Figure~\ref{fig:sim_image} shows examples of the locomotion behaviors generated by the NMPC expert in simulation. For each gait, we generated multiple trajectories with varying velocity commands to construct the training datasets. Each trajectory consists of 30,000 samples recorded at a frequency of 1 kHz, resulting in a total of 600,000 samples per gait. An additional held-out trajectory with unseen velocity inputs was used for evaluating generalization.
    
    Unlike prior work that incorporates explicit phase variable to encode the progression of motion \cite{khadiv2023learning}, our policy relies solely on raw proprioceptive inputs—namely IMU data, joint encoders, and foot contact indicators—resulting in a simpler and more practical architecture for real-world deployment.

    \subsubsection{Single Task Neural Network}
    As a baseline, we implement a simple approach by combining all gait datasets and training a single end-to-end policy without explicit task conditioning. The network architecture comprises 34 input features, three fully connected layers of 2560 neurons each, and 12 outputs corresponding to desired joint angles, as shown in Fig.~\ref{fig:NNArch}. This architecture is inspired by Nguyen et al. \cite{nguyen2021analysis}, who showed that wide networks improve learning accuracy in sensor-based control.

     \begin{figure}[t]
        \centering
        \scalebox{0.85}{  
        \begin{tikzpicture}[node distance=1.5cm]

            \tikzstyle{input neuron}=[circle, draw=black, fill=blue!20, minimum size=8mm]
            \tikzstyle{hidden neuron}=[circle, draw=black, fill=green!30, minimum size=8mm]
            \tikzstyle{output neuron}=[circle, draw=black, fill=red!30, minimum size=8mm]
        
            \tikzstyle{connector}=[->, thick]
        
            \node[input neuron] (I-1) at (-1, 3) {};
            \node at (-1, 4.0) {Input};  
         
            \node[hidden neuron] (H-1) at (0.5, 3) {2560n};
            \node at (0.5, 4.0) {Hidden 1};  
            
            \node[hidden neuron] (H-2) at (2.0, 3) {2560n};
            \node at (2.0, 4.0) {Hidden 2};  
            
            \node[hidden neuron] (H-3) at (3.5, 3) {2560n};
            \node at (3.5, 4.0) {Hidden 3};  

            \node[output neuron] (O-1) at (5.0, 3) {};
            \node at (5, 4.0) {Output};  
        
            \draw[connector] (I-1) -- (H-1);
            \draw[connector] (H-1) -- (H-2);
            \draw[connector] (H-2) -- (H-3);
            \draw[connector] (H-3) -- (O-1);
    
        \end{tikzpicture}
        }
        \caption{Baseline single‐task neural network architecture. Raw proprioceptive inputs are processed through three fully connected hidden layers to predict the desired joint position targets.}
        \label{fig:NNArch}
    \end{figure}
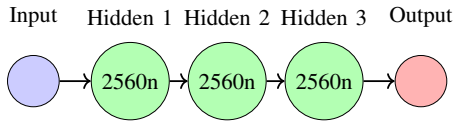
    \begin{figure}[!t]
        \centering
        \includegraphics[width=0.98\linewidth]{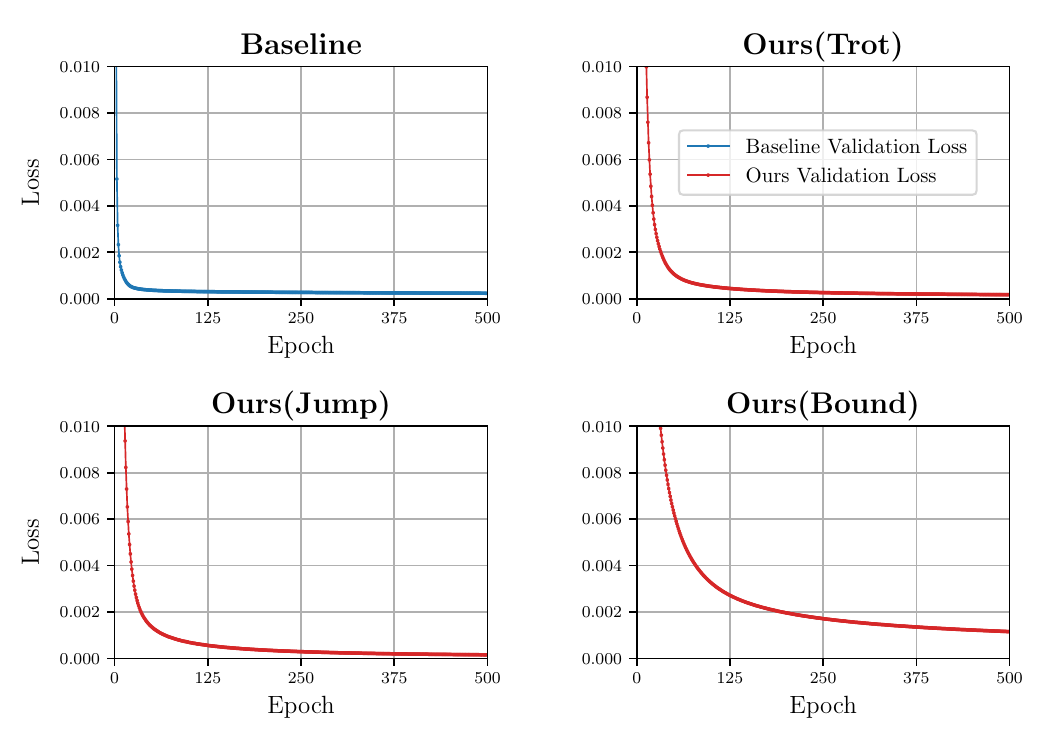}
        \vspace{-5mm}
        \caption{Validation loss curves comparing the baseline approach with our MTL method across locomotion tasks: trot, jump, and bound. Each subplot shows the validation loss over training epochs.}
        \label{fig:val_loss_comparison}
    \end{figure}

    Similar to the MTL architecture, we use the ELU activation function in the baseline. Although the baseline can learn to imitate expert behavior, it often fails to differentiate between gaits, resulting in reduced performance when switching between behaviors.
    
    \subsubsection{Evaluation Results}
    To assess performance, we compare our MTL policy against the single-task baseline on a held-out dataset containing unseen velocity combinations. As illustrated in Fig.~\ref {fig:val_loss_comparison}, both models achieve low validation loss across trot, jump, and bound gaits, suggesting that sharing a common network trunk does not compromise learning quality and can even enhance training consistency across tasks. Quantitative metrics are presented in Table~\ref{tab:eval_metrics_combined}, reporting the Mean Squared Error (MSE), Mean Absolute Error (MAE), and \( R^2 \) scores of the predicted joint target angles for each gait. Our MTL model consistently outperforms the baseline, achieving lower prediction errors and higher \( R^2 \) scores across all tasks.
    
\begin{figure*}[!t]
        \centering
        \vspace{-5mm}
        \includegraphics[width=0.8\linewidth]{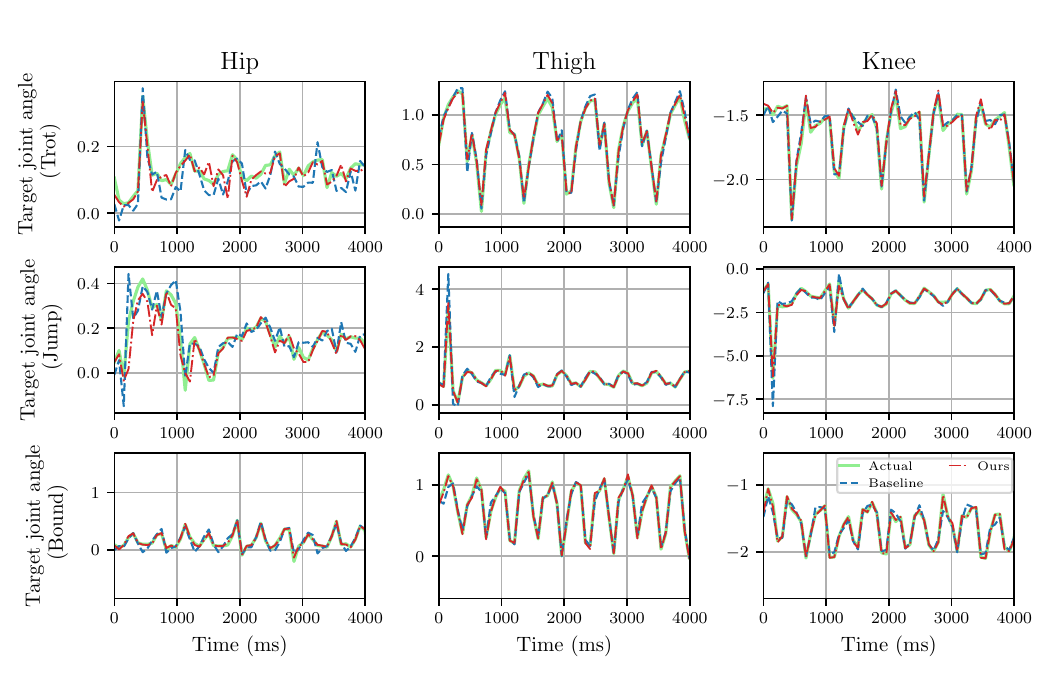}
        \vspace{-5mm}
        \caption{Predicted vs.\ expert joint target angles for the front‐left hip, thigh, and knee across trot, jump, and bound gaits in simulation. The single‐task baseline exhibits noticeable errors, whereas our MTL model closely matches the expert trajectories.}
        \label{fig:all_tasks_comparison}
    \end{figure*}

    Finally, Figure \ref{fig:all_tasks_comparison} shows a comparison between predicted and expert joint targets for the front-left leg across all three gaits. The close alignment of predicted trajectories with the expert values confirms that our MTL model accurately reproduces NMPC behavior, while the baseline model exhibits noticeable deviations.

    \begin{table}[t]
        \centering
        \caption{Evaluation metrics for predicted joint target angles across gaits in simulation. Our MTL model achieves higher \( R^2 \) scores across all tasks.}
        \label{tab:eval_metrics_combined}
        \begin{tabular}{|p{0.10\linewidth}|p{0.25\linewidth}|p{0.13\linewidth}|p{0.13\linewidth}|p{0.14\linewidth}|}
            \hline
            \cellcolor{gray!10!white} \textbf{Task} & 
            \cellcolor{gray!10!white} \textbf{Training Type} & 
            \cellcolor{gray!10!white} \textbf{MSE} \si{(\radian)\squared} & 
            \cellcolor{gray!10!white} \textbf{MAE} (\si{\radian}) & 
            \cellcolor{gray!10!white} \textbf{\(R^2\) Score} \\
            \hline
            Trot  & Single-Task & 0.001484 & 0.030213 & 0.770130 \\
                  & Multi-Task (Ours) & 0.000528 & 0.017643 & \textbf{0.919537} \\
      
            \hline
            Jump  & Single-Task & 0.006990 & 0.034837 & 0.796875 \\
                  & Multi-Task (Ours) & 0.003534 & 0.016015 & \textbf{0.892613} \\
            
            \hline
            Bound & Single-Task & 0.005179 & 0.048135 & 0.906929 \\
                  & Multi-Task (Ours) & 0.000416 & 0.011603 & \textbf{0.991763} \\

            \hline
        \end{tabular}
    \end{table}
    
    \begin{figure}[!b]
        \centering
        \includegraphics[width=0.98\linewidth]{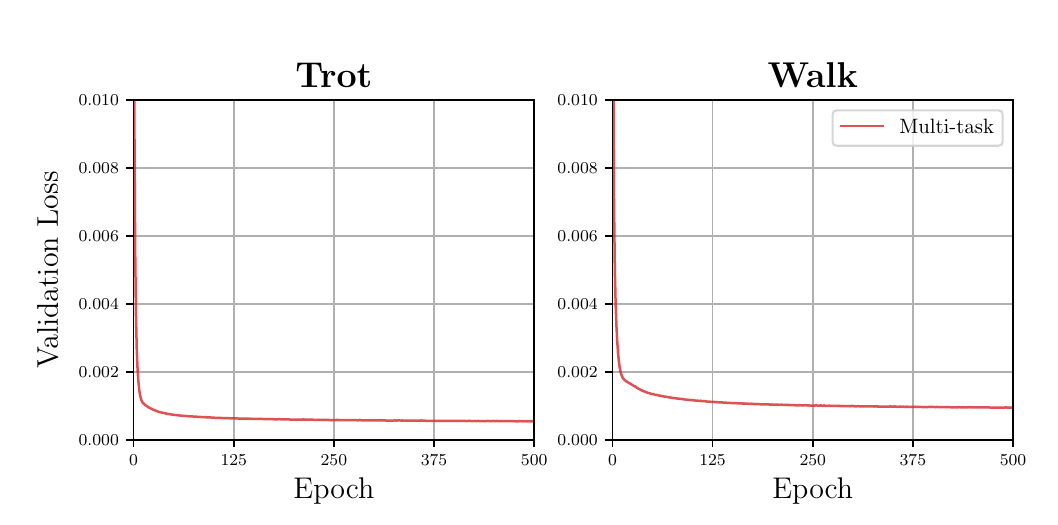}
        \vspace{-5mm}
        \caption{Validation loss curves for our multi‑task policy trained on real hardware, demonstrating stable convergence consistent with the simulation results.}
        \label{fig:val_loss_robot}
    \end{figure}
    In addition to accuracy, our unified MTL policy supports seamless gait switching at runtime. A demonstration of real-time transitions between behaviors using the same network is included in the supplementary video.

    \subsection{Real-World Experiment}
    \subsubsection{Go1 Robot Setup}
    \begin{figure*}[!t]
        \centering
        \includegraphics[width=0.8\linewidth]{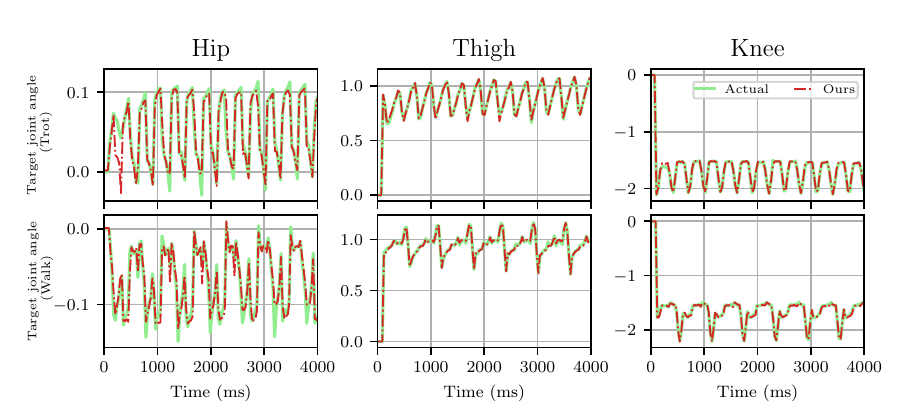}
        \vspace{-5mm}
        \caption{Predicted vs.\ expert joint target angles for trot and walk gaits on the front‑left leg of the Go1 robot in real hardware experiments. Our MTL policy closely matches the expert trajectories, demonstrating accurate reproduction of learned behaviors.}
        \label{fig:all_tasks_comparison_robot}
    \end{figure*}
    We deployed our approach on the Unitree Go1 quadruped using the legged control framework \cite{leggedcontrol, liao2023walking}, which integrates an NMPC \cite{Grandia2023mpc, Sleiman_nmpc_expensive} with Whole-Body Control (WBC). In this setup, NMPC runs at 100 Hz and WBC at 500 Hz, enabling real-time torque computation for dynamic locomotion. We executed trot and walk gaits using this controller to collect training data. Given the promising results in simulation, we adopted the same multi-task architecture and retrained it on hardware-collected data for real-world evaluation.

    \subsubsection{Evaluation Results}
    Figure~\ref{fig:val_loss_robot} presents the validation loss curves for the trot and walk gaits executed on the Go1 robot. Both behaviors exhibit smooth and consistent convergence, suggesting that the multi-task policy can be effectively trained on hardware data and remains stable during learning.

    Table~\ref{tab:eval_metrics_robot_mtl} summarizes the evaluation metrics—MSE, MAE, and $R^2$ score—of the predicted joint target angles for the multi-task network on the real robot. Consistent with the simulation results, the lower prediction errors and higher \( R^2 \) scores indicate that our MTL architecture maintains high predictive accuracy under real-world conditions, successfully replicating the expert behavior.

    \begin{table}[!ht]
        \centering
        \caption{Evaluation metrics for predicted joint target angles across gaits for our MTL model on real hardware.}
        \label{tab:eval_metrics_robot_mtl}
        \begin{tabular}{|p{0.15\linewidth}|p{0.2\linewidth}|p{0.2\linewidth}|p{0.2\linewidth}|}
            \hline
            \cellcolor{gray!10!white} \textbf{Task} & 
            \cellcolor{gray!10!white} \textbf{MSE} \si{(\radian\squared)} & 
            \cellcolor{gray!10!white} \textbf{MAE} \si{(\radian)} & 
            \cellcolor{gray!10!white} \textbf{\(R^2\) Score} \\
            \hline
            Trot  & 0.000225 & 0.007600 & 0.971015 \\
            Walk  & 0.000374 & 0.010179 & 0.961206 \\
            \hline
        \end{tabular}
    \end{table}

    Finally, Figure~\ref{fig:all_tasks_comparison_robot} presents the predicted versus expert joint targets for the front-left leg across the trot and walk gaits. The close alignment between the curves indicates that our multi-task policy, trained on real robot data, accurately replicates expert behavior in hardware experiments, confirming its validity beyond the simulation setting.

\section{Conclusion} \label{sec:conclusion}
In this work, we proposed a multi-task imitation learning framework for quadrupedal locomotion that maps raw proprioceptive sensor inputs directly to joint position targets in an end-to-end fashion. Using expert demonstrations generated by a nonlinear model predictive control policy, we trained a unified neural network in a structured PD-based action space, eliminating the need for explicit state estimation or separate task-specific controllers. Our multi-task learning architecture enables a single model to perform multiple gaits while maintaining high accuracy across tasks. We validated our approach through extensive experiments on the Unitree Go1 robot, both in simulation and real-world scenarios, demonstrating accurate reproduction of expert behavior with high \( R^2 \) scores for predicted joint target angles, and smooth gait switching.

Despite its strong performance on trained behaviors, the model did not generalize to unseen gaits such as gallop or pace, revealing limitations of standard supervised multi-task learning approaches. Future work will explore meta-learning techniques for generalization to new locomotion behaviors and investigate integrating perception modules to leverage the pre-trained locomotion policy for higher-level tasks such as navigation and environment-aware planning.

\bibliographystyle{IEEEtran}
\bibliography{references}

\end{document}